\documentclass[runningheads]{llncs}
\usepackage[T1]{fontenc}
\usepackage{graphicx}
\usepackage{booktabs}
\usepackage[misc]{ifsym}

\usepackage{mwe}

\usepackage{cite}
\usepackage{comment}
\usepackage{hyperref}
\usepackage{subfig}
\usepackage{amsmath,amssymb,amsfonts}

\usepackage{algorithm}
\usepackage{algorithmic}

\begin{document}

\title{SolarGPT-QA: A Domain-Adaptive Large Language Model for Educational Question Answering in Space Weather and Heliophysics}

\titlerunning{SolarGPT-QA}


\author{Santosh Chapagain\and
MohammadReza EskandariNasab\and
Onur Vural\and
Shah Muhammad Hamdi\and
Soukaina Filali Boubrahimi
}



\institute{Utah State University, Logan, Utah, USA \email{\{santosh.chapagain, reza.eskandarinasab, onur.vural, s.hamdi, soukaina.boubrahimi\}@usu.edu}
}


\maketitle              

\begin{abstract}
Solar activity, including solar flares, coronal mass ejections (CMEs), and geomagnetic storms can significantly impact satellites, aviation, power grids, data centers, and space missions. Extreme solar events can cause substantial economic damage with limited advance warning, underscoring the importance of early warning systems, accurate forecasting, and effective education in space science. Although large language models (LLMs) perform well on general tasks, they often lack domain specific knowledge and pedagogical capability to clearly explain complex space science concepts. We introduce SolarGPT-QA, a question answering system based on a domain adapted large language model built on the LLaMA-3 base model. The model is trained using scientific literature and large scale question and answer data generated with GPT-4 and refined using Grok-3 in a student friendly storytelling style. To evaluate response quality, we employ an \textit{LLM-as-judge} evaluation framework, where a strong reference model assesses generated answers using structured criteria including scientific accuracy, clarity, completeness, and pedagogical effectiveness. Results show that SolarGPT-QA performs strongly relative to general purpose models in zero shot settings and achieves competitive performance compared to instruction tuned models for educational explanations in space weather and heliophysics. Ablation studies indicate that combining domain adaptive pretraining with fine tuning is important for balancing scientific accuracy and educational effectiveness.

\end{abstract}

\section{Introduction}
Heliophysics studies the Sun and its interactions with the surrounding heliosphere, including the processes that drive space weather and influence Earth's technological environment \cite{nasa_heliophysics}. Within this broader heliophysical context, space weather science investigates the interconnected and multiscale interactions among the Sun, the solar wind, interplanetary space, and planetary environments, and examines how solar activity influences Earth’s space environment and technological systems \cite{schwenn2006space}. Solar activity can trigger geomagnetic storms \cite{gonzalez1994geomagnetic}. Key drivers include solar flares, coronal mass ejections (CMEs), and solar energetic particles (SEPs). These disturbances propagate through the near-Earth environment and can disrupt satellite operations, communication networks, power grids, and data centers. They can also expose astronauts and high-altitude aviation to elevated radiation hazards. For instance, the infamous \textit{Carrington
Event} in 1859 \cite{muller2014carrington} was the most intense geomagnetic storm on record, triggered by an extreme solar flare and CME, while the Quebec blackout in 1989 vividly demonstrated the susceptibility of modern technological systems to space weather disturbances. Such disruptions have resulted in economic losses of billions of dollars, and an event of similar scale today could cause even greater societal and technological consequences in our technology-driven society \cite{blong2021four}. 

Large language models have become powerful tools for understanding and explaining complex information, with growing impact across many scientific fields \cite{bharathi2024analysis}. They can answer questions, summarize research, and support learning through textual explanations. However, their use in space weather and heliophysics is still limited, largely because these areas rely on highly specialized literature and concepts that are rarely covered in general training data. One key challenge is that most space weather knowledge exclusively resides in journal articles, technical papers, and mission reports, which means general models often miss key terms, cause–effect links, or physical limits. Another issue is that teaching space science, particularly to K–12 students or early undergraduates, requires clear language, analogies, and a storytelling approach. General models are not built to balance accuracy that is tailored to the appropriate age. 

\begin{figure}[htbp]
\centerline{\includegraphics[width=0.9\linewidth]{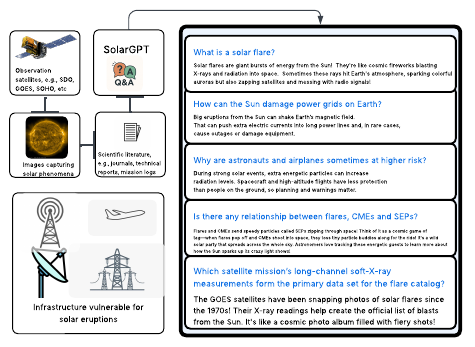}}
\caption{Overview of the SolarGPT-QA system illustrating its educational question–answering framework for space weather and heliophysics.}
\label{fig:qa}
\end{figure}

To fill this gap, we introduce SolarGPT-QA, a domain-adapted model designed for educational question answering in space weather and heliophysics for K–12 learners. Fig.~\ref{fig:qa} presents an overview of the SolarGPT-QA educational question–answering framework. The model is trained on a carefully curated corpus of high-impact, peer-reviewed research articles published since 2010, aligned with the Solar Dynamics Observatory (SDO) era of modern flare and CME research. These sources are further refined for relevance to solar eruptions and used to generate a large collection of synthetic question–answer pairs. Rather than targeting expert-level analysis, SolarGPT-QA focuses on helping students grasp core concepts through clear, accessible explanations to questions such as how solar flares form or why the Sun can disrupt technology on Earth. 

\section{Related Work}

\subsection{Forecasting of Solar Flares, CMEs, and SEPs}
Existing forecasting approaches can be broadly categorized into empirical, physics-based, and data-driven methods \cite{georgoulis2024prediction}. Empirical techniques exploit correlations between historical proxies, such as sunspot magnetic complexity and X-ray flux, and subsequent eruptions \cite{lin2003theories}. Physics-based models rely on magnetohydrodynamic simulations and coronal magnetic field extrapolations \cite{wang2013mhd}, but their operational use is limited by computational cost and incomplete boundary conditions \cite{whitman2023sepmodels}. Furthermore, although significant progress has been made in understanding solar magnetic field dynamics, a quantitative theory that rigorously connects the accumulation and restructuring of magnetic flux to the onset of extreme solar events remains lacking in physics-driven approaches, leaving the key mechanisms governing flare and coronal mass ejection initiation still not fully understood \cite{ahmadzadeh2021train}. This gap highlights the importance of data-driven approaches and machine learning methods for improving the prediction and characterization of extreme solar events. The launch of NASA’s Solar Dynamics Observatory (SDO) in 2010 marked a major advance in this area by providing continuous, high-resolution observations of the Sun. Data-driven machine learning approaches have shown substantial promise, particularly for flare and SEP prediction. Multivariate time series models capture the temporal evolution of solar active regions and have demonstrated improved predictive skill over traditional threshold-based methods. Hamdi et al. \cite{hamdi2017timeseries} formulated solar flare prediction as an end-to-end time series classification problem using photospheric magnetic field parameters. Progress in this area has been supported by the development of benchmark datasets. The Space Weather Analytics for Solar Flares (SWAN-SF) multivariate time series dataset integrates magnetic field measurements with flare labels and has become a widely used resource for solar flare prediction research \cite{angryk2020swansf}. Machine learning-based models in this domain include classical approaches such as logistic regression \cite{song2009statistical}, support vector machines \cite{bobra2015solar}, and decision trees \cite{ma2017decisiontrees}, as well as deep learning architectures for temporal sequence modeling, including LSTMs \cite{muzaheed2021sequence}, contrastive learning approaches, graph-based methods, and transformer frameworks \cite{vural2024contrastive, vural2025global}. Subsequent work addressed severe class imbalance and data scarcity, demonstrating that preprocessing, sampling strategies, and data augmentation significantly improve robustness for rare, high-impact events \cite{eskandarinasab2024preprocessing}. Many efforts have applied deep learning on solar imagery, including magnetograms and extreme ultraviolet observations, to capture spatial precursors to flares and CMEs, while multimodal fusion of time series and images further enhances forecasting accuracy \cite{park2018cnn,hosseinzadeh2025multimodal}. Despite these advances, most operational systems predict flares, CMEs, and SEPs independently and provide limited explanatory context, functioning largely as black boxes \cite{collado2019ccmc,petrenko2023ccmc,petrenko2024barriers}. These limitations motivate unified, multimodal, and explainable forecasting frameworks that build on prior advances in time series modeling, dataset development, and interpretability.

\subsection{LLMs for Scientific and Technical Domains}
LLMs have evolved rapidly in recent years and can achieve state-of-the-art performance across a wide range of tasks and model scales \cite{brown2020language, Chapagain2025AdvancingMinority}. Subsequent work has explored the scaling behavior of LLMs, architectural variations, and expanded application domains, demonstrating the versatility of LLMs in both general-purpose and specialized settings \cite{rae2021scaling,zhang2022opt}. DAPT and fine-tuning are common techniques for improving LLMs for specific datasets or domains. Similarly, retrieval-augmented generation (RAG) has been widely used to improve factual grounding by integrating external knowledge sources into language models \cite{lewis2020rag}. In parallel, instruction tuning has emerged as a key alignment technique for adapting LLMs to follow user intent and task-specific objectives \cite{wei2022finetuned}. Beyond general-purpose models, large language models have increasingly been adapted to specialized scientific domains, including medicine, biology, Earth science, and environmental sciences. In the medical domain, Med-PaLM \cite{singhal2022large} demonstrates that large-scale language models can encode clinical knowledge and perform competitively on medical question answering through instruction tuning and expert evaluation. Generalist medical foundation models, such as those proposed by Moor et al. \cite{moor2023med}, extend this approach by enabling a single model to support a wide range of clinical reasoning tasks. MedGPT \cite{kraljevic2021medgpt} leverages electronic health records and named entity recognition to predict future medical events, while BioGPT \cite{luo2022biogpt} is pre-trained on large-scale biomedical literature to improve biomedical text generation and knowledge extraction. Geneformer \cite{theodoris2023geneformer} is a life science domain-adapted language model that applies transformer-based language modeling to single-cell transcriptomic data, enabling transfer learning for gene network inference and cellular state prediction. Similarly, large-scale protein language models demonstrate that language modeling techniques can predict atomic-level protein structures directly from amino acid sequences \cite{lin2023protein}. More recently, scientific LLMs have been introduced for Earth, ocean, and geoscience applications. GeoGPT \cite{deng2023geoscience} represents one of the first large language models explicitly designed for geoscience tasks. OceanGPT \cite{bi2024oceangpt} is trained on marine and oceanographic literature, demonstrating superior performance and preliminary embodied intelligence for diverse ocean science and engineering tasks compared to general-purpose LLMs. CLLMate \cite{li2025cllmate} focuses on climate-related applications, integrating climate science literature and data-driven knowledge to support climate analysis, interpretation, and decision support. Tele-Knowledge Pretraining \cite{chen2023tele} incorporates structured technical knowledge into language models for fault diagnosis and engineering analysis. Despite these advances, most existing scientific LLMs primarily target expert-level analysis and professional decision-making, often producing technically dense and jargon-heavy outputs. In contrast, our work focuses on domain-adapted language modeling for educational scientific question answering, emphasizing conceptual clarity, accessibility, and age-appropriate explanations. Rather than supporting expert users, SolarGPT-QA is explicitly designed to facilitate learning and understanding in complex scientific domains such as space weather and heliophysics.

\section{SolarGPT-QA}
\begin{figure}[htbp]
\centerline{\includegraphics[width=0.8\linewidth]{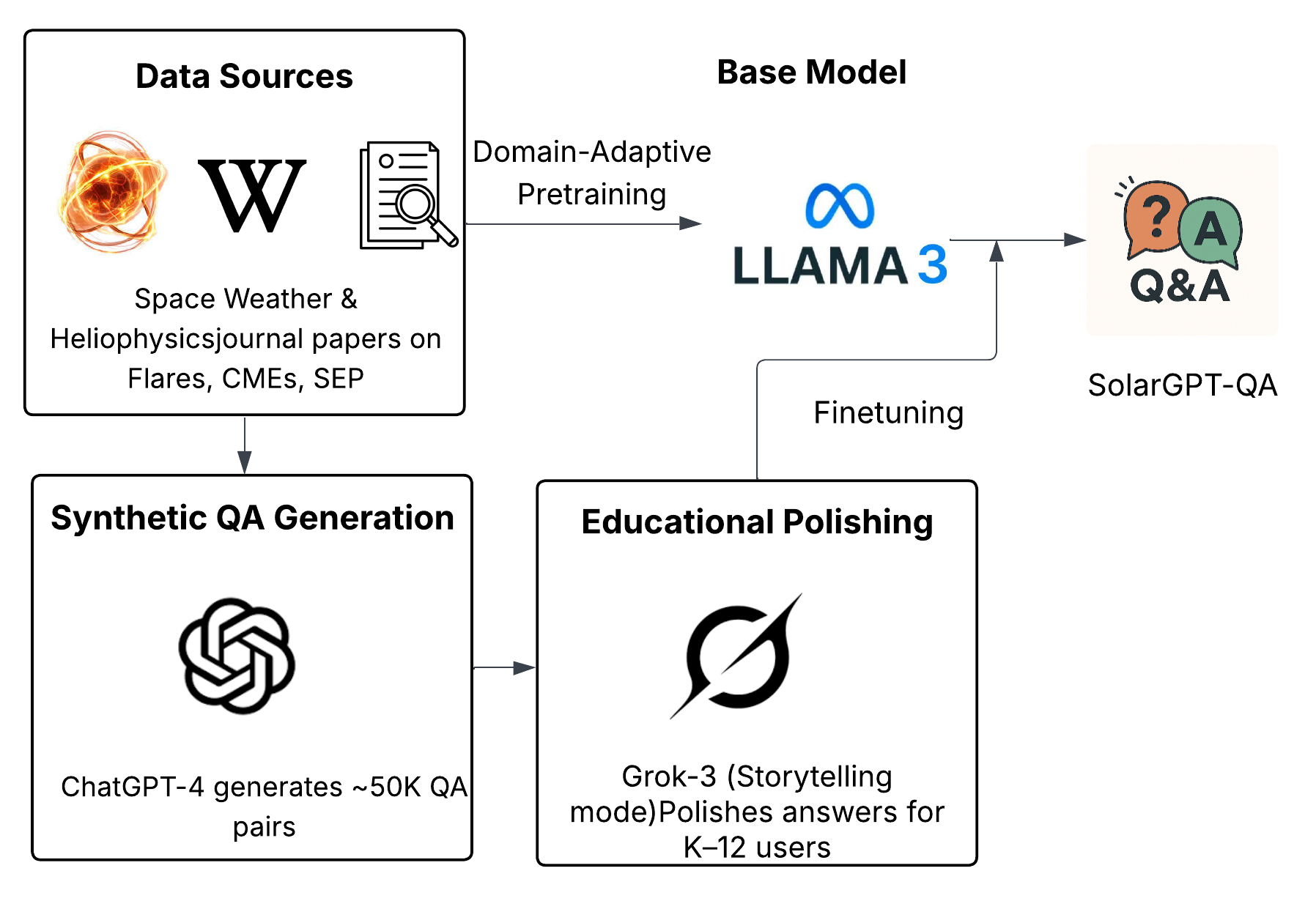}}
\caption{System architecture and workflow of SolarGPT-QA, showing the data sources, domain adaptation, and question–answer generation pipeline.}
\label{fig:SoalarGPT}
\end{figure}

SolarGPT-QA is a domain-adapted large language model designed specifically for educational question answering in space weather and heliophysics, with an emphasis on K–12–friendly explanations. The methodology consists of four main stages: (1) domain corpus construction and preprocessing, (2) domain-adaptive pretraining (DAPT) using LoRA, (3) educational question–answer fine-tuning, and (4) inference-time constraints to promote clarity and reduce hallucinations. An overview of the system architecture is shown in Fig.~\ref{fig:SoalarGPT}. Let $\mathcal{D}_H$ denote a corpus of heliophysics literature, and $\mathcal{D}_{QA}$ denote a supervised QA dataset designed for K--12 learners. Our objective is to learn a conditional language model
\begin{equation}
p_\theta(y \mid x),
\end{equation}
where $x$ is a heliophysics question, and $y$ is an age-appropriate explanatory answer that preserves scientific correctness while prioritizing conceptual clarity.

\subsection{Domain Corpus Construction and Normalization}
We curate a heliophysics corpus $\mathcal{D}_H = \{d_i\}_{i=1}^N$ from peer-reviewed literature published between 2010 and 2024, during the SDO operational period. Each document $d_i$ is extracted from PDF format and transformed using a deterministic preprocessing function:
\begin{equation}
\tilde{d}_i = f_{\text{clean}}(d_i),
\end{equation}
where $f_{\text{clean}}$ removes references, citations, section headers, figure captions, URLs, and formatting artifacts, followed by Unicode normalization.

The cleaned corpus $\mathcal{T} = \{\tilde{d}_i\}$ is tokenized into sequences
\begin{equation}
\mathbf{x} = (x_1, \dots, x_T), \quad x_t \in \mathcal{V},
\end{equation}
where $\mathcal{V}$ denotes the tokenizer vocabulary of the base language model.

\subsection{Domain-Adaptive Pretraining with Low-Rank Adaptation}

We initialize SolarGPT-QA from a pretrained causal language model, Meta-LLaMA-3-8B, and adapt it to the heliophysics domain using \emph{domain-adaptive pretraining (DAPT)} \cite{gururangan2020don} combined with \emph{Low-Rank Adaptation (LoRA)} \cite{hu2022lora} for parameter efficiency. As defined in Section~3.1, the heliophysics corpus is processed as
\[
\mathcal{D}_H \rightarrow \mathcal{T}
\]

DAPT minimizes the autoregressive negative log-likelihood over the tokenized corpus $\mathcal{T}$:
\begin{equation}
\mathcal{L}_{\text{DAPT}}(\theta) =
\mathbb{E}_{x \sim \mathcal{T}}
\left[
\sum_{t=1}^{T}
-\log p_\theta(x_t \mid x_{<t})
\right]
\end{equation}

This objective biases the model toward heliophysics-specific terminology, causal structures, and semantic patterns associated with solar flares, coronal mass ejections (CMEs), and solar energetic particles (SEPs).

For a transformer weight matrix, $W \in \mathbb{R}^{d \times k}$, LoRA applies a low-rank update
\begin{equation}
W \rightarrow W' = W + \Delta W,
\end{equation}
Here, $\Delta W$ is the learned adaptation applied to the frozen pretrained weight matrix $W$. LoRA parameterizes this update as the product of two low-rank matrices, $\Delta W = BA$, where $A \in \mathbb{R}^{r \times k}$ projects the original representation into a lower-dimensional space and $B \in \mathbb{R}^{d \times r}$ projects it back to the original dimension. The rank $r \ll \min(d,k)$ controls the number of trainable parameters and enables parameter-efficient adaptation. The overall training workflow is summarized in Algorithm 1, and the LoRA configuration values are provided in Section~4.

\subsection{Fine-Tuning on QA dataset}

We define a supervised dataset
\begin{equation}
\mathcal{D}_{QA} = \{(q_i, a_i)\}_{i=1}^M,
\end{equation}
where each answer $a_i$ is rewritten to emphasize analogies, simplified causal explanations, and age-appropriate language.

Fine-tuning minimizes the conditional likelihood:
\begin{equation}
\mathcal{L}_{QA}(\theta) =
\mathbb{E}_{(q,a) \sim \mathcal{D}_{QA}}
\left[
-\log p_\theta(a \mid q)
\right].
\end{equation}

During this stage, all base model parameters remain frozen, and only LoRA parameters are updated. Inputs are formatted explicitly as:
\begin{center}
\texttt{Q: <question> \\ A: <answer>}
\end{center}
to align the model with educational discourse patterns.

\begin{algorithm}[t]
\caption{Training Procedure for \textsc{SolarGPT-QA}}
\label{alg:solargpt}
\begin{algorithmic}[1]
\STATE \textbf{Require:} Heliophysics corpus $\mathcal{D}_H$, QA dataset $\mathcal{D}_{QA}$, pretrained model $p_{\theta_0}$
\STATE Clean and normalize $\mathcal{D}_H \rightarrow \mathcal{T}$
\STATE Initialize LoRA adapters on attention layers
\STATE \textbf{Domain-Adaptive Pretraining:}
\FOR{each batch $\mathbf{x} \in \mathcal{T}$}
    \STATE Update LoRA parameters using $\mathcal{L}_{\text{DAPT}}$
\ENDFOR
\STATE \textbf{Educational Fine-Tuning:}
\FOR{each $(q,a) \in \mathcal{D}_{QA}$}
    \STATE Format input as \texttt{Q: q A: a}
    \STATE Update LoRA parameters using $\mathcal{L}_{QA}$
\ENDFOR
\RETURN Trained \textsc{SolarGPT-QA} model
\end{algorithmic}
\end{algorithm}

The overall training objective of \textsc{SolarGPT-QA} can be summarized as:
\begin{equation}
\mathcal{L} = \mathcal{L}_{\text{DAPT}} + \lambda \mathcal{L}_{QA},
\end{equation}
where $\mathcal{L}_{\text{DAPT}}$ denotes the domain-adaptive pretraining loss and 
$\mathcal{L}_{QA}$ denotes the educational question--answer fine-tuning loss, with 
$\lambda$ controlling the emphasis on pedagogical alignment. 
In practice, these objectives are optimized \emph{sequentially rather than jointly}: 
the model is first adapted to heliophysics literature using $\mathcal{L}_{\text{DAPT}}$, 
and the resulting DAPT-adapted checkpoint is subsequently fine-tuned using 
$\mathcal{L}_{QA}$ to produce the final \textsc{SolarGPT-QA} model. 
The combined objective summarizes the cumulative learning goal across training stages, 
rather than a single joint optimization. Unlike prior scientific language models optimized for expert-level analysis, 
\textsc{SolarGPT-QA} is explicitly designed for educational deployment. 
By combining domain-adaptive pretraining, parameter-efficient optimization, 
and pedagogically aligned fine-tuning with constrained decoding, the model bridges 
expert heliophysics knowledge and accessible K--12 instruction.

\subsection{Inference time Hallucination Mitigation}
In addition to domain-adaptive pretraining and educational fine-tuning, we applied several inference time controls to make responses more dependable. These methods do not modify the model’s parameters; instead, they guide decoding so that answers remain short, focused, and well formed. We used controlled sampling ($T=0.7$, top-$k=50$, top-$p=0.9$) with a repetition penalty of 1.1 to reduce repetitive continuations. We also restricted response length with a 128-token limit and a custom stopping criterion that ends generation at the first newline. Finally, we filtered the generated text to retain only complete sentences, removing unfinished fragments that could otherwise introduce vague content. These steps help limit hallucinations and improve answer clarity, as reflected in the ablation study in Section 4.6, where the models without inference time hallucination mitigation showed weaker results.

\section{Experiments}
\subsection{Training Details and Compute}
\label{sec:training_details}
We conducted both DAPT and QA fine-tuning on Meta-LLaMA-3-8B using Low-Rank Adaptation (LoRA). To reduce memory usage, the pretrained LLaMA-3-8B backbone was loaded in 8-bit precision, while training computations for the LoRA parameters were performed in FP16 (16-bit floating point). Rather than updating all model parameters, LoRA adapters were inserted into the query and value projection layers of the attention modules (i.e., \texttt{q\_proj} and \texttt{v\_proj}), with rank $r{=}16$, scaling $\alpha{=}32$, and dropout $p{=}0.05$. Training used the AdamW optimizer with a learning rate of $2\times10^{-4}$, linear warmup of 100 steps, and a maximum sequence length of 512 tokens. The per-device batch size was 1 with gradient accumulation of 8, resulting in an effective batch size of 8 sequences per optimization step. Both DAPT and QA fine-tuning were run for 3 epochs. All experiments were conducted on a Linux server with dual Intel Xeon Gold 5220R CPUs (24 cores each, 2.20~GHz) and four NVIDIA RTX A5000 GPUs (24~GB VRAM). Individual training runs used a single GPU, with GPU selection controlled at runtime. To support reproducibility and transparent reporting of compute usage, our training scripts automatically log the total number of optimization steps, wall-clock training time, and peak GPU memory consumption to a \texttt{training\_log.json} file. The total number of training tokens can be derived from these logs as: $ \text{tokens} \approx (\text{total steps}) \times 512 \times 8$, corresponding to the sequence length and effective batch size used during training. All of our codes and datasets used in this experiment are available on GitHub\footnote{\url{https://github.com/hlabcsusu/SolarGPT}}.

\subsection{Dataset}

The datasets used in this study consist of two components: 
(1) a heliophysics domain corpus for domain-adaptive pretraining (DAPT) and 
(2) a question--answer (QA) dataset for educational fine-tuning and evaluation.

\subsubsection{Domain Corpus for DAPT}
\begin{figure}[htbp]
\centerline{\includegraphics[width=0.9\linewidth]{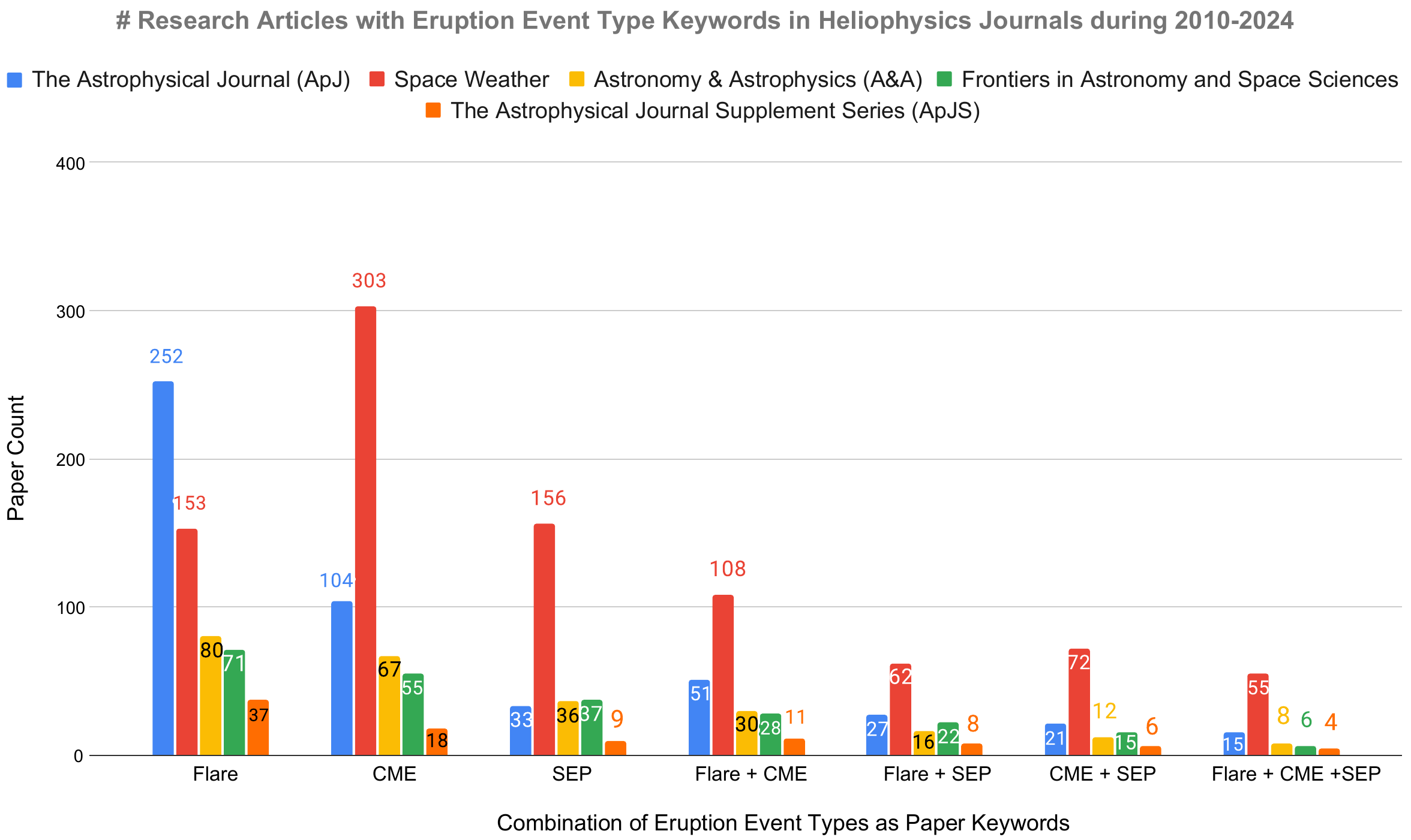}}
\caption{Combination of Eruption Event Types as Paper Keywords}
\label{fig:papstat}
\end{figure}

\begin{figure}[htbp]
\centerline{\includegraphics[width=0.7\linewidth]{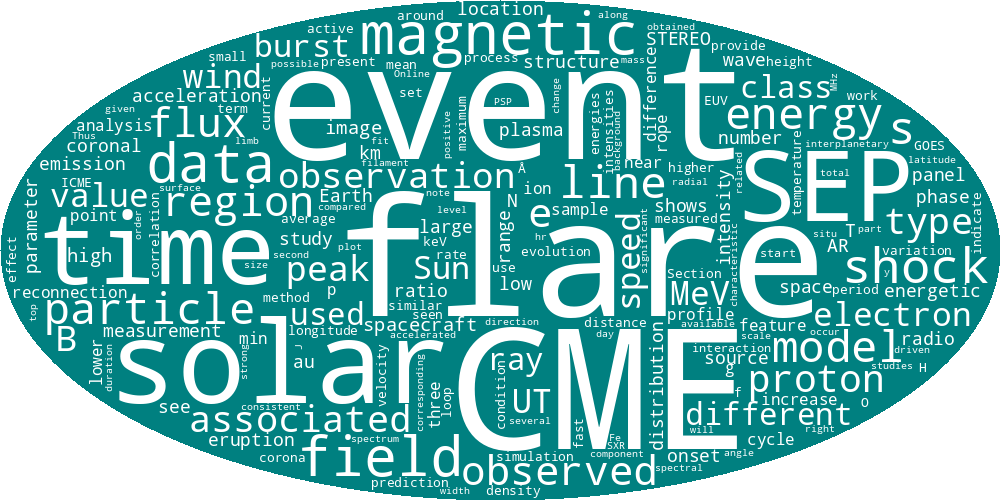}}
\caption{WordCloud of the data corpus used for DAPT}
\label{fig:wc1}
\end{figure}

The heliophysics corpus is curated from peer-reviewed research articles published between 2010 and 2024, corresponding to the operational era of the SDO. Articles were selected from high-impact journals with an h5-index greater than 40 to ensure scientific rigor and relevance. These venues include \textit{The Astrophysical Journal (ApJ)}, \textit{The Astrophysical Journal Supplement Series (ApJS)}, \textit{Astronomy \& Astrophysics (A\&A)}, \textit{Space Weather}, and \textit{Frontiers in Astronomy and Space Sciences}, among others. Papers were filtered using eruption-related keyword co-occurrence such as \textit{solar flare}, \textit{coronal mass ejection (CME)}, and \textit{solar energetic particles (SEP)}. Fig.~\ref{fig:papstat} shows the distribution of eruption-related research articles across five leading heliophysics journals between 2010 and 2024, stratified by individual and combined event types. This analysis identifies 1,570 relevant papers from these venues alone, and extending the selection to all qualifying journals yields an estimated 5,966 research articles used for domain adaptation. All documents used were obtained from publicly accessible sources, including open-access journal articles and author-provided PDFs available through publisher websites or institutional repositories. No proprietary databases or restricted-access collections were used. The corpus was used solely for internal model training and was not redistributed in raw or processed form, in accordance with publisher terms of use. All documents were extracted from PDF format and cleaned by removing references, citations, figure captions, URLs, and formatting artifacts. Corpus-level statistics confirm strong coverage of eruption-related concepts. In particular, the word cloud shown in Fig.~\ref{fig:wc1} highlights the dominance of key heliophysics terms such as \textit{solar}, \textit{CME}, \textit{flare}, \textit{SEP}, \textit{magnetic field}, \textit{particle}, and \textit{shock}, indicating that the corpus effectively captures the semantic structure and terminology of solar eruptive phenomena.

\subsubsection{QA Dataset Construction}
The supervised QA dataset consists of approximately 5,000 question--answer pairs derived from curated heliophysics literature using a human-in-the-loop approach, where initial LLM-generated candidates were rigorously validated by domain experts for scientific accuracy and K--12 pedagogical alignment. To ensure a robust assessment, a separate test set of 300 QA pairs was extracted from PDF documents that were not included in the training corpus, effectively eliminating information leakage. This test set is stratified into three difficulty levels: \textit{easy} questions cover basic definitions and solar phenomena for elementary learners; \textit{medium} questions require reasoning about physical processes like CME interactions with the magnetosphere; and \textit{hard} questions demand the synthesis of complex concepts, such as the relationship between flares and SEPs. This tiered structure enables a systematic evaluation of SolarGPT-QA across varying levels of conceptual depth and learner maturity while maintaining consistent scientific grounding.

\subsection{Baseline}
We evaluate SolarGPT-QA against several strong general-purpose large language models that are widely used for scientific question answering and education. The baselines include ChatGPT (GPT-4o), Grok (Grok-1.5), Claude (Claude 3 Sonnet), DeepSeek (DeepSeek-V2), and Gemini (Gemini 1.5 Pro). All baseline models are evaluated under a zero-shot setting using identical prompts and decoding parameters without retrieval augmentation, to ensure a fair and controlled comparison. In addition, we include an instruction-tuned variant of ChatGPT, where the model is prompt-tuned using a small number of representative examples (few-shot setting), in order to compare SolarGPT-QA against a highly optimized general-purpose model with explicit pedagogical alignment.

\subsection{Evaluation Method}
To evaluate our model, we employ an \textit{LLM-as-a-Judge} \cite{gu2024survey} framework, utilizing a \textit{GPT-4o} to conduct automated pairwise assessments in place of manual human review. We test model responses against a stratified set of 300 questions consisting of  100 easy, 150 medium, and 50 hard test instances in which outputs from SolarGPT-QA and a baseline model are presented to the evaluator in a randomized, anonymized order to eliminate positional bias. The judge assesses each pair based on a structured rubric tailored for K–12 heliophysics education, specifically measuring factual correctness, clarity of explanation, age appropriateness, and educational usefulness. Each comparison results in a win, tie, or loss for SolarGPT-QA, with the final win rate calculated as the proportion of preferred responses. We repeat each comparison multiple times and the final win rate is reported as the mean across runs with standard deviation.

\subsection{Results}
\begin{figure}[htbp]
\centerline{\includegraphics[width=1\linewidth]{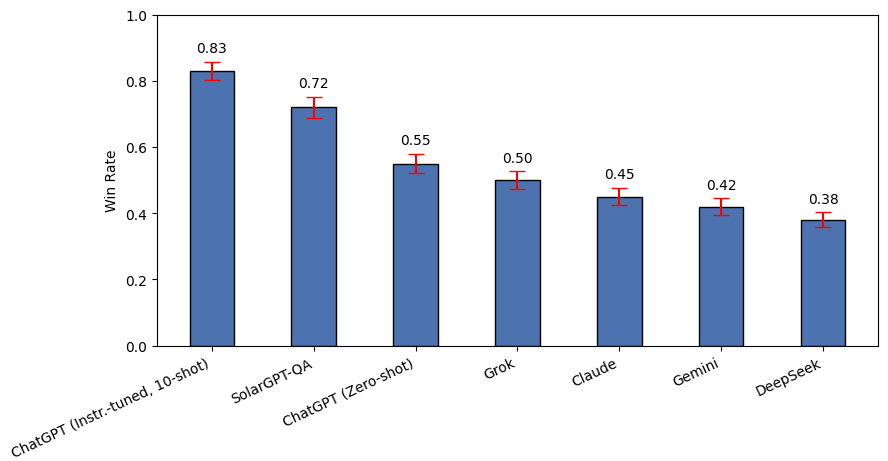}}
\caption{\textit{LLM-as-a-Judge} win-rate comparison across models. Bars represent the mean win rate over repeated evaluation runs, and error bars denote the standard deviation. Numerical labels above each bar indicate the mean win rate.}
\label{fig:winrate}
\end{figure}

We evaluate SolarGPT-QA against several strong general-purpose large language
models using LLM-based pairwise comparison. Fig.~\ref{fig:winrate} presents
the win-rate comparison across models. ChatGPT (instruction-tuned, 10-shot)
achieves the highest win rate at 83\%, indicating that the judge model
consistently prefers its responses. SolarGPT-QA achieves a win rate of 72\%,
outperforming all zero-shot baselines, including ChatGPT (55\%), Grok (50\%),
Claude (45\%), Gemini (42\%), and DeepSeek (38\%). These results suggest that
domain-adaptive pretraining combined with educational fine-tuning substantially
improves response quality in specialized scientific question answering, while
remaining competitive with strongly prompted general-purpose models.

\subsection{Ablation Study}
We conduct an ablation study to analyze the impact of domain-adaptive pretraining (DAPT) and educational QA fine-tuning in SolarGPT-QA. Four model variants are compared: the base LLaMA-3 model, LLaMA-3 with DAPT only, LLaMA-3 with QA fine-tuning only, and the full SolarGPT-QA model. DAPT significantly improves scientific grounding and the correct use of heliophysics terminology. However, responses produced by the DAPT-only model often contain explanations that are overly technical for K--12 learners. In contrast, QA fine-tuning alone improves readability and explanatory structure but sometimes lacks sufficient domain depth, leading to incomplete or simplified explanations. Table~\ref{tab:ablation} shows that the full SolarGPT-QA model achieves the highest win rate, indicating that combining domain-adaptive pretraining with educational fine-tuning yields the strongest overall performance. The ablation results indicate that both components contribute complementary benefits. Domain-adaptive pretraining strengthens scientific accuracy and terminology usage, while QA fine-tuning improves explanation quality and pedagogical clarity. Their combination produces responses that better balance scientific rigor with educational accessibility.

\begin{table}[t]
\centering
\caption{Ablation study evaluating DAPT and QA fine-tuning.}
\label{tab:ablation}
\setlength{\tabcolsep}{3pt}
\begin{tabular}{lc}
\hline
\textbf{Model Variant} & \textbf{Win Rate (\%, mean $\pm$ std)} \\
\hline
Base LLaMA-3-8B & $28.0 \pm 1.9$ \\
LLaMA-3 + DAPT  & $54.0 \pm 2.4$ \\
LLaMA-3 + QA Fine-Tuning & $48.0 \pm 2.2$ \\
\textbf{SolarGPT-QA} & $\mathbf{72.0 \pm 3.2}$ \\
\hline
\end{tabular}
\end{table}

\section{Discussion and Conclusion}

\subsection{Discussion}
The results demonstrate that domain adaptation alone is insufficient for effective educational explanations and that pedagogical alignment is an important factor when deploying large language models for science education. While general-purpose LLMs such as ChatGPT and Grok exhibit strong linguistic fluency and factual recall, they often produce explanations that are either overly technical or insufficiently structured for K--12 learners in specialized domains such as heliophysics. Using an \textit{LLM-as-a-Judge} evaluation framework, we find that instruction-tuned ChatGPT (10-shot) achieves the highest overall win rate, indicating strong performance when explicit instructional alignment is introduced through prompting. SolarGPT-QA performs competitively and outperforms all zero-shot general-purpose baselines, achieving a 72\% win rate. This suggests that combining domain-adaptive pretraining with educational fine-tuning can significantly improve the quality of explanations in specialized scientific domains without requiring few-shot prompting during inference. The ablation study further clarifies the contribution of each training stage. Domain-adaptive pretraining improves scientific grounding and terminology usage but tends to preserve expert-level language. QA fine-tuning improves readability and explanatory structure but may lack domain depth when used alone. The full SolarGPT-QA model offers the best balance, giving explanations that are both scientifically accurate and easy to understand. More broadly, these findings suggest that many existing scientific LLMs are optimized for expert-level reasoning rather than educational communication. As AI systems increasingly support science outreach and education, integrating domain knowledge with pedagogical alignment will be essential for producing explanations that are both accurate and accessible.

\subsection{Limitations}
Both training and evaluation QA pairs were initially generated using large language models. Although strict document-level separation between training and test sources is enforced, residual stylistic artifacts from synthetic generation may remain. Additionally, the evaluation relies on an \textit{LLM-as-a-Judge} framework, which may introduce bias associated with the judge model's own preferences or training distribution. While pairwise evaluation and randomized response ordering help mitigate these effects, further validation using independent evaluation methods remains important for future work.

\subsection{Future Work}
This study represents an initial step toward domain-adapted language models for heliophysics education. Future work will focus on developing \textit{SolarGPT-Core}, a larger domain-adapted foundation model trained on an expanded corpus of heliophysics literature to improve scientific depth and coverage of rare solar events. We also plan to develop \textit{SolarGPT-PredEx}, a multimodal model that integrates solar imagery and time-series measurements with language modeling to generate both space weather predictions and natural language explanations grounded in observational data. Additional improvements to SolarGPT-QA will explore retrieval-augmented generation (RAG), curriculum-aware prompting, and adaptive difficulty control to support classroom and outreach applications.

\subsection{Conclusion}
We introduced SolarGPT-QA, a domain-adapted large language model designed for educational question answering in space weather and heliophysics. By combining domain-adaptive pretraining on heliophysics literature with QA fine-tuning, which improves the clarity and accessibility of scientific explanations. Evaluation using an \textit{LLM-as-a-Judge} framework shows that SolarGPT-QA outperforms multiple zero-shot general-purpose models and approaches the performance of strongly instruction-tuned baselines. Ablation studies further demonstrate that combining domain adaptation with educational fine-tuning is important for balancing scientific accuracy with pedagogical clarity. Overall, this work provides a practical approach for building domain-specific educational language models and represents an initial step toward a broader SolarGPT ecosystem that integrates scientific knowledge, explanation, and education in heliophysics.

\begin{credits}
\subsubsection{\ackname} Shah Muhammad Hamdi is supported by the GEO directorate under NSF awards \#2301397 and \#2530946. Soukaina Filali Boubrahimi is supported by GEO Directorate under NSF awards \#2204363, \#2240022, and \#2530946.
\end{credits}

%
%
%
%

\bibliography{source.bib} 
\bibliographystyle{splncs04}

\end{document}